\documentclass[a4paper,fleqn]{cas-dc}
\usepackage[numbers,sort&compress]{natbib}
\usepackage{balance}

\def\tsc#1{\csdef{#1}{\textsc{\lowercase{#1}}\xspace}}
\tsc{WGM}
\tsc{QE}

\begin{document}
\let\WriteBookmarks\relax

\shorttitle{ECSdiff}    

\shortauthors{B. Vasic and B. Vasic}  

\title [mode = title]{Energy-Conditioned Noise Schedule and Whitening for Spectral Diffusion}  

\tnotemark[1] 

\tnotetext[1]{Bane Vasic acknowledges the support of the National Science Foundation under grant CCF-2100013, and a generous gift from his  friends and Maecenases Dora and Barry Bursey.} 

\author[1,2]{Bata Vasic}[bioid=1, orcid = 0000-0001-8667-1507]

\cormark[1]
\ead{bata.vasic@elfak.ni.ac.rs}
\credit{Conceptualization of this study, Methodology, Software}
\affiliation[1]{organization={University of Nis, Faculty of Electronic Engineering},
            addressline={Aleksandra Medvedeva 4}, 
            city={Nis},
            postcode={18104}, 
            state={},
            country={Serbia}}
\affiliation[2]{organization={University MB, Information Technology department},
            addressline={Teodora Drajzera 27}, 
            city={Belgrade},
            postcode={11040}, 
            state={},
            country={Serbia}}
            
\author[3]{Bane Vasic}[bioid=2, orcid = 0000-0003-2365-4106]
\ead{vasic@ece.arizona.edu}
\credit{Methodology, Supervision}
\affiliation[3]{organization={University of Arizona, Electrical \& Computer Engineering (ECE) department},
            addressline={1230 E. Speedway Blvd.}, 
            city={Tucson},
            postcode={85721-0104}, 
            state={AZ},
            country={USA}}

\cortext[1]{Corresponding author}


\begin{abstract}
This paper introduces an energy-adaptive noise scheduling and whitening strategy for transform-domain diffusion models. Existing spectral diffusion methods account for the non-uniform statistics of transform coefficients through coefficient scaling, normalization, or frequency prioritization, while the forward diffusion noise schedule remains largely independent of the underlying spectral-energy distribution. We investigate whether the temporal evolution of the forward diffusion process should also follow the spectral organization of natural images. The proposed  formulation combines global spectral whitening with energy-conditioned noise allocation that jointly modulates the injected noise according to the energy of individual transform coefficients and an image-dependent energy path over diffusion time. The resulting forward process preserves Gaussian transitions with closed-form marginals and remains compatible with standard DDPM and DDIM procedures without modifying the diffusion architecture. Experiments on CIFAR-10 demonstrate the contribution of the proposed energy-conditioned noise schedule and spectral whitening, reducing Fréchet Inception Distance from 142.48 for a compact DCTdiff U-Net variant to 100.45.
\end{abstract}


\maketitle


\section{Introduction}
\label{introduction}
Denoising diffusion models~\cite{sohl2015, ho2020, song2022} formulate image generation as the reverse of a stochastic process that progressively transforms data into Gaussian noise. Denoising Diffusion Probabilistic Models (DDPM)~\cite{ho2020} and Denoising Diffusion Implicit Models (DDIM)~\cite{song2022} provide stochastic and deterministic realizations of this principle, respectively, while continuous-time formulations have been investigated in~\cite{ysong2020}. Most existing diffusion models operate directly on pixel representations, where the forward diffusion process applies an identical noise schedule to every image coordinate.

Recently, several studies have investigated diffusion in transform-domain representations, including the Fourier~\cite{crabb2024,falck2025} and discrete cosine transform (DCT)~\cite{ahmed1974} domains~\cite{ning2024,gao2024}. Unlike pixel representations, transform coefficients possess well-defined spectral meaning and exhibit highly non-uniform statistical properties. In natural images, most of the signal energy is concentrated in a relatively small number of low-frequency DCT coefficients, whereas higher-frequency components contain substantially less energy. Consequently, the transform domain is characterized by a pronounced imbalance between spectral components.

Recent transform-domain diffusion models explicitly account for this imbalance. DCTDiff~\cite{ning2024} introduces coefficient scaling, signal-to-noise ratio (SNR) scaling, and frequency prioritization to improve optimization in the DCT domain. Moreover, the authors show that image diffusion exhibits a spectral autoregressive behavior, indicating that transform coefficients contribute unequally throughout the diffusion process. These observations suggest that spectral diffusion fundamentally differs from pixel-domain diffusion and should exploit the underlying spectral organization of natural images.

Motivated by these observations, we investigate a complementary question that, to the best of our knowledge, has not been addressed previously: should the forward diffusion noise schedule itself depend on the spectral-energy distribution? Existing transform-domain diffusion methods adapt the representation, coefficient scaling, or optimization strategy, whereas the temporal evolution of the forward diffusion process remains largely independent of the spectral energy associated with individual transform coefficients.

This paper introduces an \emph{energy-conditioned noise scheduling and whitening} strategy for transform-domain diffusion models. Rather than proposing a new diffusion architecture, the objective is to investigate spectral-energy-aware noise scheduling as an independent component of existing transform-domain diffusion frameworks. The proposed schedule depends on diffusion time, coefficient energy, and image-dependent spectral energy, assigning smaller perturbations to dominant spectral components while modulating their temporal evolution according to the spectral-energy distribution of each image. The resulting formulation preserves the standard Gaussian diffusion framework and remains directly compatible with conventional DDPM and DDIM training and sampling procedures.

To isolate the contribution of the proposed scheduling strategy, the experimental evaluation employs the same U-Net architecture and identical training configuration with both uniform and energy-conditioned spectral noise schedules. The objective is not to compare different diffusion architectures, but to evaluate the influence of energy-conditioned scheduling within the same transform-domain diffusion framework. The proposed formulation is presented in Section~\ref{sec:method}, its relation to DCTdiff is discussed in Section~\ref{sec:SOTA}, and the experimental evaluation is given in Section~\ref{sec:experiment}.

\section{Baseline Spectral Diffusion Framework}
\label{sec:framework}

This section introduces the baseline transform-domain diffusion framework adopted throughout this work. Following the JPEG coding pipeline~\cite{wallace1992}  and the block-DCT representation introduced in DCTdiff~\cite{ning2024}, the image is converted into compact spectral coefficients before diffusion.

Unlike DCTdiff, which employs a Transformer backbone, the present work uses the resulting spectral representation as the input to a conventional U-Net diffusion model. This baseline serves as the reference framework upon which the proposed Energy-Conditioned Spectral Diffusion (ECSdiff) is introduced in the next section.

\subsection{Block-DCT Representation}

Let $x\in\mathbb{R}^{H\times W\times3}$ denote an RGB image. Following the JPEG image coding pipeline, the image is first transformed into the luminance--chrominance color space, $x'=\mathbf{M}x+\mathbf{b}$, where $\mathbf{M}$ is the RGB-to-YCbCr transformation matrix and $\mathbf{b}$ is the corresponding offset vector. Since the human visual system is less sensitive to chrominance than luminance, $2\times$ chroma subsampling is applied to the $C_b$ and $C_r$ channels, yielding one luminance channel and two subsampled chrominance channels.

Each channel is subsequently partitioned into non-overlapping $B\times B$ blocks. Let $A\in\mathbb{R}^{B\times B}$ denote one image block and let $D(u,v)$ denote its spectral coefficient at spatial frequency $(u,v)$. The two-dimensional DCT~\cite{ahmed1974} is
\begin{equation}
D(u,v)=
\sum_{x=0}^{B-1}
\sum_{y=0}^{B-1}
A(x,y)\,
\phi_u(x)\,
\phi_v(y),
\label{eq-dct}
\end{equation}
\noindent where $\phi_u(x)=\alpha(u)\cos[\pi(2x+1)u/(2B)]$, $\phi_v(y)=\alpha(v)\cos[\pi(2y+1)v/(2B)]$, and $\alpha(q)=\sqrt{1/B}$ for $q=0$ and $\alpha(q)=\sqrt{2/B}$ for $q>0$.

\subsection{Compact Spectral Representation}

The coefficients of each DCT block are reordered according to the JPEG zig-zag scan, producing a one-dimensional sequence arranged from low to high spatial frequencies. Four neighboring luminance blocks together with the corresponding chrominance blocks are grouped into a single $4Y+1Cb+1Cr$ spectral unit, preserving the spatial correspondence between luminance and chrominance information.

Since high-frequency coefficients contribute little to the reconstructed image quality, the last $m^{*}$ coefficients are removed from every zig-zag sequence following~\cite{ning2024}, and determined as
\begin{equation}
m^*=
\arg\max_m
\left\{
m:
\operatorname{FID}
(P_{\mathrm{data}},P_{\mathrm{DCT}}(m))
<
\eta
\right\},
\label{eq-mstar}
\end{equation}
\noindent where $P_{\mathrm{DCT}}(m)$ denotes the image distribution reconstructed after removing the last $m$ high-frequency coefficients, $\eta$ is a predefined threshold, and $\operatorname{FID}(\cdot,\cdot)$ denotes the Fréchet Inception Distance (FID)~\cite{heusel2017}. 

Let $\mathbf{x}_0$ denote the resulting compact spectral representation, containing $K=B^2-m^*$ retained DCT coefficients per block, used throughout the remainder of this work.

\subsection{Baseline Spectral Diffusion}

The compact spectral representation $\mathbf{x}_0$ is modeled using the Markov process~\cite{noris1997} and standard DDPM formulation with a uniform forward noise schedule~\cite{ho2020},
\begin{equation}
q(\mathbf{x}_t|\mathbf{x}_0)=
\mathcal{N}
\left(
\sqrt{\bar{\alpha}_t}\,\mathbf{x}_0,
(1-\bar{\alpha}_t)\mathbf{I}
\right),
\label{eq-ddpm}
\end{equation}
\noindent where $\bar{\alpha}_t=\prod_{i=1}^{t}(1-\beta_i)$.

A denoising network $\varepsilon_\theta(\mathbf{x}_t,t)$ is trained by minimizing
\begin{equation}
\mathcal{L}
=
\mathbb{E}_{\mathbf{x}_0,\varepsilon,t}
\left[
\left\|
\varepsilon-
\varepsilon_\theta(\mathbf{x}_t,t)
\right\|_2^2
\right].
\label{eq-loss}
\end{equation}

\subsection{Reverse Sampling}

Sampling starts from $\mathbf{x}_T\sim\mathcal{N}(\mathbf{0},\mathbf{I})$ and follows the standard DDPM or DDIM reverse process. After denoising, the retained coefficients are restored to the block-DCT representation, the discarded high-frequency coefficients are set to zero, and the inverse block DCT reconstructs the YCbCr image, which is finally converted back to the RGB color space.

\section{Energy-Conditioned Spectral Diffusion}
\label{sec:method}

The baseline framework described in the previous section follows the standard DDPM formulation with a uniform forward noise schedule. Recent theoretical analysis of transform-domain diffusion has shown that the diffusion process inherently exhibits spectral autoregressive behavior, whereby high-frequency components progressively lose their signal-to-noise ratio earlier than low-frequency components under isotropic Gaussian perturbations~\cite{ning2024}. This behavior follows directly from the power-law energy profile of natural images and provides a theoretical justification for treating individual spectral components differently during diffusion.

Rather than modifying the diffusion architecture, the proposed ECSdiff explicitly incorporates this observation into the forward process by introducing an energy-conditioned noise schedule. The proposed formulation consists of three consecutive stages. First, the compact spectral representation is whitened to reduce the statistical imbalance between spectral components. Second, the spectral-energy distribution is estimated from the original transform coefficients. Finally, the forward noise schedule is modulated according to the estimated spectral energy.

\subsection{Spectral Whitening}

Let $\mathbf{x}_0=[x_1,\ldots,x_K]^\top$ denote the compact spectral representation obtained in Section~\ref{sec:framework}. Although the Block-DCT representation reduces the statistical mismatch between the pixel and transform domains, the retained coefficients still exhibit differences in variance across spectral indices. To reduce this imbalance, each coefficient is normalized independently according to
\begin{equation}
\tilde{x}_k=
\omega_k
\frac{x_k}{s_k},
\qquad
s_k=
\sqrt{\mathbb{E}[x_k^2]+\delta_s},
\label{eq-norm}
\end{equation}
\noindent where $\mathbb{E}[\cdot]$ denotes the empirical expectation over the training set, $\delta_s>0$ prevents division by zero, and $\omega_k$ is a frequency-dependent weighting factor. This dataset-level normalization treats $s_k^2$ as the ensemble energy scale of the $k$-th spectral component, equalizing systematic differences between spectral modes while preserving image-specific energy variations \cite{vanhemmen1980,nvidia2024}.

The weighting factor combines a diffusion-step adaptation and an epoch-dependent adaptation, $\omega_k=
\omega_t(k)\omega_e^{\lambda_e}(k)$, where $\lambda_e\ge0$ controls the contribution of the epoch-adaptive term. 
For a spectral component located at frequency $(u,v)$, the diffusion-step weighting is defined as
\begin{equation}
\begin{aligned}
\omega_t(u,v)
=
\omega_{\mathrm{lo}}(t)
+
\left(
\omega_{\mathrm{hi}}(t)
-
\omega_{\mathrm{lo}}(t)
\right)r(u,v),\\
r(u,v) =
\sqrt{u^2+v^2}/
\sqrt{(B-1)^2+(B-1)^2}.
\end{aligned}
\label{eq-white}
\end{equation}
\noindent where $r(u,v)$ is the normalized radial frequency. The functions $\omega_{\mathrm{lo}}(t)$ and $\omega_{\mathrm{hi}}(t)$ are defined as $\omega_{\mathrm{lo}}(t)=
\omega_{\mathrm{lo}}^{(0)}+(\omega_{\mathrm{lo}}^{(1)}-\omega_{\mathrm{lo}}^{(0)})\tau_t$ and $\omega_{\mathrm{hi}}(t)=
\omega_{\mathrm{hi}}^{(0)}+(\omega_{\mathrm{hi}}^{(1)}-\omega_{\mathrm{hi}}^{(0)})\tau_t$, with $\tau_t=(t/(T-1))^{\lambda_t}$, where $\lambda_t>0$ controls the temporal progression.

The epoch-adaptive weighting follows the same formulation with $\tau_e=(e/(N_e-1))^{p_e}$, where $e$ and $N_e$ denote the current and total number of training epochs, respectively. The resulting normalized coefficients constitute the representation
$\tilde{\mathbf{x}}^{(i)}=[\tilde{x}_1^{(i)},\ldots,\tilde{x}_K^{(i)}]^\top$,
for $i=1,\ldots,N$, used as input to the proposed diffusion model.

\subsection{Spectral Energy Estimation}

The energy-conditioned schedule is determined from the original spectral coefficients rather than from the whitened representation. Let $E_k(0)=\mathbb{E}[x_k^2]$ denote the empirical energy of the $k$-th spectral coefficient over the training set. After logarithmic energy compression, the corresponding initial normalized spectral-energy distribution is defined as
\begin{equation}
\rho_k(0)=
\frac{\log\left(1+\lambda E_k(0)\right)}
{\sum_{j=1}^{K}\log\left(1+\lambda E_j(0)\right)+\delta},
\label{eq-SE}
\end{equation}
\noindent where $\delta>0$ ensures numerical stability and $\lambda$ controls the logarithmic compression. The distribution $\rho_k(0)$ characterizes the relative energy of each spectral component and provides the global spectral-energy distribution for constructing the coefficient-dependent noise adaptation.

\subsection{Energy-Conditioned Noise Schedule}

The spectral-energy distribution defined in \eqref{eq-SE} is used to construct the coefficient-dependent energy-adaptation factor
\begin{equation}
\psi_k=
\frac{K
\left(
\rho_k(0)+\delta
\right)^{-\gamma}}
{\sum_{j=1}^{K}
\left(
\rho_j(0)+\delta
\right)^{-\gamma}},
\label{eq-psi}
\end{equation}
\noindent where $\gamma\ge0$ controls the strength of energy adaptation.

For each training image $i$, let $\rho_k^{(i)}$ denote its normalized spectral-energy distribution computed from the original spectral coefficients. The corresponding cumulative spectral-energy distribution is defined as
\begin{equation}
R_k^{(i)}=
\frac{\sum_{j=1}^{k}\rho_j^{(i)}}
{\sum_{j=1}^{K}\rho_j^{(i)}+\delta_E}.
\label{eq-Rk}
\end{equation}

The corresponding energy-dependent temporal modulation is initially defined as
\begin{equation}
\tilde{\beta}_E^{(i)}(t)=\beta_1+(\beta_T-\beta_1)g^{(i)}(t),
\label{eq-betaE}
\end{equation}
\noindent where $g^{(i)}(t)$ is the normalized form of $-\log(1-R^{(i)}(t)+\delta_E)$, with $g^{(i)}(1)=0$ and $g^{(i)}(T)=1$. The dimensionless temporal modulation factor is then obtained by mean normalization,
\begin{equation}
\beta_E^{(i)}(t)=
\frac{\tilde{\beta}_E^{(i)}(t)}
{\frac{1}{T}\sum_{s=1}^{T}\tilde{\beta}_E^{(i)}(s)+\delta_E}.
\label{eq-betaE-norm}
\end{equation}

The energy-conditioned forward diffusion schedule and the corresponding retention factors are then defined as
\begin{equation}
\beta_{k,t}^{(i)}=\beta_t\beta_E^{(i)}(t)\psi_k,
\
\alpha_{k,t}^{(i)}=1-\beta_{k,t}^{(i)},
\
\bar{\alpha}_{k,t}^{(i)}
=
\prod_{s=1}^{t}\alpha_{k,s}^{(i)}.
\label{eq-alpha}
\end{equation}

The normalization of $\psi_k$ preserves its mean value across spectral components. Higher-energy components are perturbed more conservatively, while lower-energy components receive stronger perturbations, with the temporal evolution defined by $\beta_t$ and conditioned by the image-dependent factor $\beta_E^{(i)}(t)$.

\subsection{Energy-Conditioned Spectral Diffusion and Sampling}

For a training image $i$, the forward diffusion process is defined using the
image-dependent schedule from
\eqref{eq-alpha} as
\begin{equation}
x_{k,t}^{(i)}
=
\sqrt{\bar{\alpha}_{k,t}^{(i)}}\,
\tilde{x}_k^{(i)}
+
\sqrt{1-\bar{\alpha}_{k,t}^{(i)}}\,
\varepsilon_k,
\
\varepsilon_k\sim\mathcal{N}(0,1),
\label{eq-fordiff}
\end{equation}
\noindent where $\tilde{x}_k^{(i)}$ denotes the $k$-th coefficient of the whitened spectral representation $\tilde{\mathbf{x}}^{(i)}$ introduced in~\eqref{eq-norm}. The denoising network is trained using the baseline objective~\eqref{eq-loss}, with the forward process given by~\eqref{eq-fordiff}.

During sampling, an energy-dependent noise path $\beta_E^{(i)}(t)$ is first
obtained from a reference image, and the corresponding coefficient-dependent
schedule $\beta_{k,t}^{(i)}$ defined above is used throughout the reverse
process. For DDIM sampling,
\begin{equation}
x_{k,t-1}^{(i)}
=
\sqrt{\bar{\alpha}_{k,t-1}^{(i)}}\,
\hat{x}_{k,0}^{(i)}
+
\sqrt{1-\bar{\alpha}_{k,t-1}^{(i)}}\,
\varepsilon_{\theta,k}(\mathbf{x}_t^{(i)},t),
\label{eq-ddim}
\end{equation}
\noindent where $\hat{x}_{k,0}^{(i)}$ denotes the estimated clean $k$-th spectral coefficient of image $i$. Thus, the proposed ECSdiff preserves the standard DDPM and DDIM sampling formulations while using the image-dependent energy path $\beta_E^{(i)}(t)$ as conditioning for the corresponding coefficient-dependent noise schedule in the reverse process.

\section{Relation to DCTdiff}
\label{sec:SOTA}

\begin{figure}
	\centering
	\includegraphics[width=.9\columnwidth]{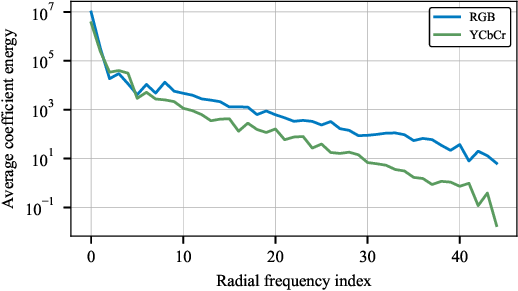}
	\caption{Radial DCT energy profile in the YCbCr and RGB domains for the full-spectrum representation.}
	\label{fig1}
\end{figure}

ECSdiff introduces a spectral diffusion strategy that advances the DCTdiff philosophy~\cite{ning2024} while retaining its baseline framework defined in Section~\ref{sec:framework}.

DCTdiff compensates for spectral imbalance through Entropy-Consistent Scaling (ECS), Entropy-Based Frequency Reweighting (EBFR), and signal-to-noise ratio (SNR) scaling. ECS applies a common factor $\eta$ to all retained coefficients, $x_k'=x_k/\eta$, whereas EBFR weights the denoising objective according to the entropy of individual frequency distributions. Since ECS moves low-energy coefficients closer to zero, DCTdiff additionally scales the SNR of the default noise schedule to compensate for their earlier corruption.

Our strategy replaces global scaling with component-wise normalization $\tilde{x}_k=x_k/s_k$~\eqref{eq-norm}. The resulting spectral statistics are further incorporated through the factors $\psi_k$~\eqref{eq-psi} and $\beta_E^{(i)}(t)$~\eqref{eq-betaE-norm}, defining the coefficient- and image-dependent schedule $\beta_{k,t}^{(i)}=\beta_t\beta_E^{(i)}(t)\psi_k$~\eqref{eq-alpha}. Hence, spectral statistics directly control normalization and noise allocation.

This formulation is consistent with the spectral autoregression established by Theorem~5.1 in~\cite{ning2024}, according to which isotropic diffusion noise first corrupts low-energy high-frequency components and progressively affects lower frequencies. DCTdiff exhibits this ordering implicitly, whereas the proposed schedule explicitly incorporates spectral energy into the forward process through $\beta_{k,t}^{(i)}$.

A further distinction arises during sampling. Since the energy-conditioned $\beta_{k,t}^{(i)}$ path is image-dependent, it is retained in the reverse process, allowing sampling to follow the spectral-energy characteristics of the conditioning image and potentially benefit conditioned image generation.

\begin{figure}
	\centering
	\includegraphics[width=.9\columnwidth]{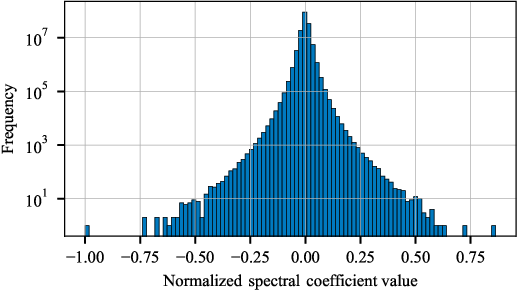}
	\caption{Profile of YCbCr spectral coefficients after normalization~\eqref{eq-norm} and spectral weighting~\eqref{eq-white} in the full-spectrum DCT representation.}
	\label{fig2}
\end{figure}

\section{Experimental Verification and Analysis}
\label{sec:experiment}

The experimental analysis considers two spectral diffusion settings: (i) full-spectrum DCT diffusion to examine the limitations of uniform spectral diffusion models (USDM), and (ii) compact block-DCT diffusion for evaluating the proposed ECSdiff and its comparison with DCTdiff~\cite{ning2024} under a common experimental setting.

\subsection{Experimental Setup}

Experiments are conducted on CIFAR-10 using the standard training and test partitions of respectively $50{,}000$ and $10{,}000$ RGB images with a resolution of $32\times32$. 

For direct comparison, the DCTdiff framework is adapted from its original transformer-based implementation to the same U-Net backbone used by ECSdiff, while preserving its spectral representation and diffusion procedure. This provides a common, lower-capacity setting for evaluating the effect of the proposed energy-conditioned modifications.

All models use $T=1000$ diffusion steps and are evaluated using both DDPM and DDIM sampling. A Python implementation of ECSdiff is available at~\cite{bata2026}.

Unless stated otherwise, ECSdiff uses $\gamma=0.2$ in~\eqref{eq-psi}, and $\delta=10^{-5}$ and $\lambda=1$ in~\eqref{eq-SE}. A moderate $\gamma$ is used because $\psi_k$ acts multiplicatively with $\beta_t$ and the image-dependent $\beta_E^{(i)}(t)$ on an already whitened representation, whereas stronger spectral weighting would amplify the combined global and image-dependent energy conditioning.

The spectral weighting~\eqref{eq-white} is defined by $\omega_{\mathrm{lo}}^{(0)}=1.0$, $\omega_{\mathrm{lo}}^{(1)}=0.8$, $\omega_{\mathrm{hi}}^{(0)}=0.01$, and $\omega_{\mathrm{hi}}^{(1)}=0.6$, with $p_e=1$, $\lambda_e=0.08$, and $\lambda_t=2.0$ over $1000$ training epochs. The learning rate follows
$lr(e)=lr_{\min}+(lr_{\max}-lr_{\min})(1-e/N_e)^{\gamma_{lr}}$,
with $lr_{\max}=4\times10^{-4}$, $lr_{\min}=4\times10^{-6}$, and $\gamma_{lr}=1.0$.

\subsection{Full-Spectrum DCT Diffusion}
\label{Experiment 1}

The first experiment evaluates direct full-spectrum DCT diffusion, without the block-DCT organization and coefficient reduction defined in~\eqref{eq-dct} and~\eqref{eq-mstar}. This setting preserves the full spectral imbalance and therefore provides a demanding case for evaluating the proposed conditioning.

The transition from the pixel domain to the full DCT representation produces a highly non-uniform spectral-energy profile dominated by low-frequency coefficients, as shown in Fig.~\ref{fig1}. Applying the ECSdiff normalization~\eqref{eq-norm} and spectral weighting~\eqref{eq-white} to the full-spectrum representation reduces this imbalance and produces a more compact coefficient profile centered around zero, as shown in Fig.~\ref{fig2}.

\subsection{Compact DCT Spectral Diffusion}
\label{Experiment 2}

The second experiment evaluates ECSdiff within the compact block-DCT framework defined in Section~\ref{sec:framework} and examines the contributions of spectral whitening and energy-conditioned noise scheduling.
\begin{figure}
	\centering
	\includegraphics[width=.9\columnwidth]{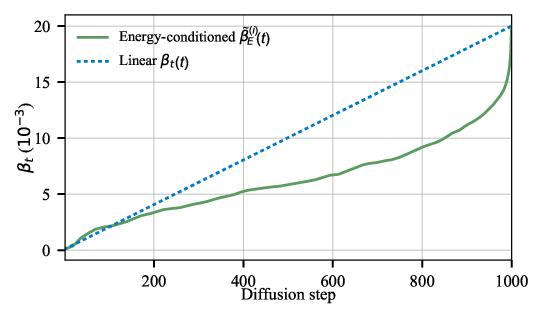}
	\caption{Example of an image-dependent energy-conditioned noise path $\tilde{\beta}_E(t)$ before mean normalization to the dimensionless modulation factor $\beta_E(t)$, compared with the linear noise schedule $\beta_t$.}
	\label{fig3}
\end{figure}

Figure~\ref{fig3} shows an example of the image-dependent energy-conditioned noise path obtained from a single image. Compared with the linear schedule, $\tilde{\beta}_E^{(i)}(t)$ introduces noise more gradually over most of the diffusion process and increases toward $\beta_T$ at the final steps. The displayed path is shown before the mean normalization in~\eqref{eq-betaE-norm}, which produces the dimensionless modulation factor $\beta_E^{(i)}(t)$ used in~\eqref{eq-alpha} as conditioning modulation during training and subsequently during sampling.
\begin{figure}
	\centering
	\includegraphics[width=.9\columnwidth]{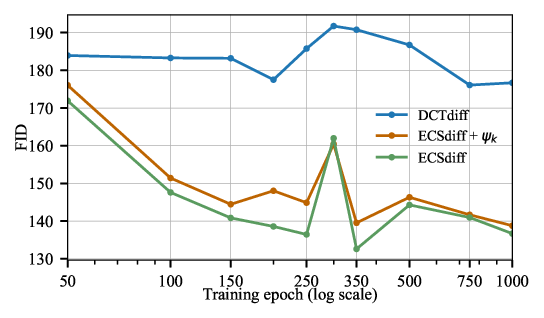}
	\caption{FID evolution of DCTdiff, ECSdiff with $\psi_k$ and ECSdiff during training, evaluated using $1000$ generated samples at each checkpoint. The training-epoch axis is shown on a logarithmic scale.}
	\label{fig4}
\end{figure}

The quantitative performance of ECSdiff is evaluated against the pixel-domain DDPM, full-spectrum USDM, and DCTdiff by monitoring FID at selected training epochs, as reported in Table~\ref{table:fid}.

\begin{table}[width=\linewidth,cols=8,pos=t]
\caption{FID evolution during training.}
\label{table:fid}
\begin{tabular*}{\tblwidth}{@{} LCCCCCCC@{}}
\toprule
\textbf{Config.} &
\textbf{50} &
\textbf{100} &
\textbf{150} &
\textbf{250} &
\textbf{350} &
\textbf{500} &
\textbf{1000} \\
\midrule
Pixel
& 107.19 & 106.81 & 100.30 & 98.23 & 98.41 & 95.26 & 86.69 \\
USDM
& 233.89 & 226.44 & 182.36 & 213.54 & 166.80 & 188.30 & 192.24 \\
DCTdiff
& 183.93 & 183.28 & 183.19 & 185.76 & 190.77 & 186.73 & 176.69 \\
\textbf{ECSdiff}
& \textbf{171.92} & \textbf{147.61} & \textbf{140.85} & \textbf{138.59}
& \textbf{136.46} & \textbf{132.61} & \textbf{136.68} \\
\bottomrule
\end{tabular*}
\footnotesize
Frechet Inception Distance (FID) values are reported at selected training
epochs using $1000$ generated samples and the same U-Net configuration.
\end{table}

In the minimal U-Net configuration, the spectral-domain models do not reach the FID of the pixel-domain DDPM. Among the spectral methods, however, ECSdiff consistently achieves the lowest FID and enters a lower-FID regime after approximately $250$ epochs, with smaller improvements thereafter. This indicates that energy-conditioned spectral diffusion can reach a stable performance regime with fewer training epochs, which may reduce computational cost when applied to higher-capacity architectures.

The difference in training behavior is further illustrated in Fig.~\ref{fig4}. ECSdiff reaches lower FID values early in training and remains below DCTdiff at all evaluated checkpoints, indicating that the improvement is established during the earlier stages of training rather than only after prolonged optimization.

To assess the dependence of the FID estimate on the number of generated samples, DCTdiff and ECSdiff models trained for $1000$ epochs were additionally evaluated using $500$, $1000$, $2500$, $5000$, and $10{,}000$ samples. The results are reported in Table~\ref{table:fid_samples}. ECSdiff consistently achieves lower FID than DCTdiff across all sample sizes. A similar dependence on sample size is expected at the remaining checkpoints, for which $1000$ samples were used to limit computational cost.

\begin{table}[width=\linewidth,cols=6,pos=t]
\caption{FID evaluation with different numbers of generated samples.}
\label{table:fid_samples}
\begin{tabular*}{\tblwidth}{@{} LCCCCC@{}}
\toprule
\textbf{Config.} &
\textbf{500} &
\textbf{1000} &
\textbf{2500} &
\textbf{5000} &
\textbf{10000} \\
\midrule
DCTdiff
& 197.84 & 176.69 & 155.11 & 148.84 & 142.48 \\
\textbf{ECSdiff}
& \textbf{157.78} & \textbf{136.68} & \textbf{114.39}
& \textbf{104.58} & \textbf{100.45} \\
\bottomrule
\end{tabular*}
\footnotesize
FID values for DCTdiff and ECSdiff trained for $1000$ epochs, evaluated
using different numbers of generated samples under the same U-Net
configuration and the same Gaussian noise seed for sampling.
\end{table}

\subsection{Ablation Analysis}

The ablation analysis isolates the contribution of spectral whitening and energy-conditioned noise scheduling within the compact spectral framework. Starting from USDM, the EAFW configuration introduces dataset-level spectral whitening, while ECSdiff additionally incorporates the image-dependent energy-conditioned noise schedule.

At $1000$ epochs, USDM achieves an FID of $192.2413$, while DCTdiff achieves $176.6944$. Introducing spectral whitening in EAFW further reduces the FID to $152.7026$, and adding the energy-conditioned noise schedule reduces it to $136.6838$ for ECSdiff. These results support the dataset-level whitening in~\eqref{eq-norm} as a mechanism for reducing systematic differences between spectral modes, while preserving image-specific energy variations used by $\beta_E^{(i)}(t)$.

\subsection{Sampling Performance}

\begin{figure*}
	\centering
	\setlength{\tabcolsep}{2pt}
	\begin{tabular}{cccc}
		\includegraphics[width=0.24\textwidth]{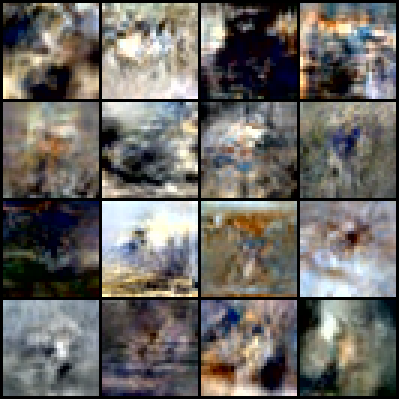} &
		\includegraphics[width=0.24\textwidth]{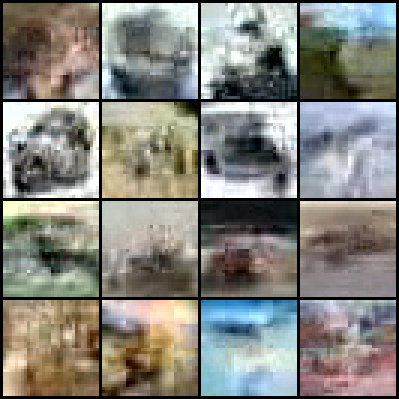} &
		\includegraphics[width=0.24\textwidth]{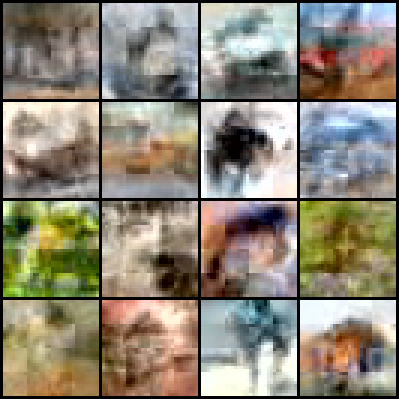} &
		\includegraphics[width=0.24\textwidth]{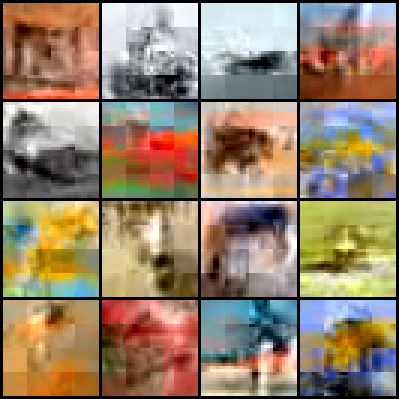} \\
		\includegraphics[width=0.24\textwidth]{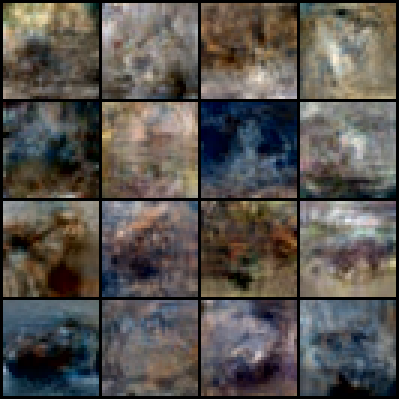} &
		\includegraphics[width=0.24\textwidth]{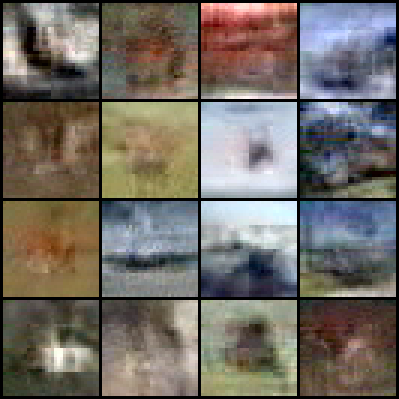} &
		\includegraphics[width=0.24\textwidth]{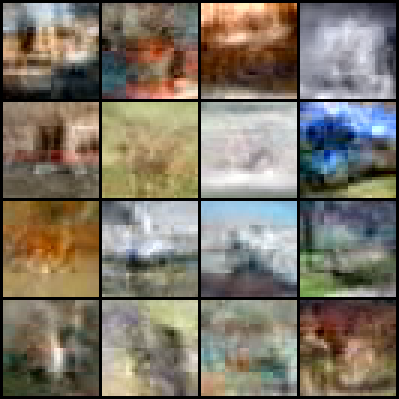} &
		\includegraphics[width=0.24\textwidth]{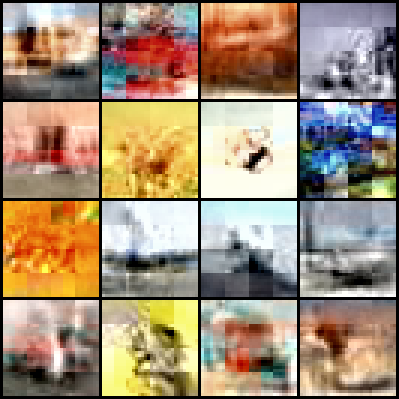} \\
		USDM & DCTdiff & EAFW & ECSdiff
	\end{tabular}
	\caption{Sampling comparison after $1000$ training epochs using the same seed from the test set for all methods. From left to right: USDM, DCTdiff, EAFW, and ECSdiff. Upper row: DDPM sampling. Lower row: DDIM sampling~\eqref{eq-ddim}.}
	\label{fig5}
\end{figure*}

The sampling performance is evaluated using both DDPM and DDIM to examine the visual behavior of the spectral diffusion models after $1000$ training epochs. Figure~\ref{fig5} compares the generated samples under the same compact representation and training configuration.

Figure~\ref{fig5} shows the sampling behavior of USDM and DCTdiff and isolates the effect of energy-conditioned diffusion by comparing EAFW and ECSdiff. ECSdiff produces coherent object structures under both DDPM and DDIM sampling. Compared with EAFW, the main differences are observed in spectral amplitude and color contrast rather than object structure. The corresponding FID results show improved distributional similarity for ECSdiff, while the visual comparison indicates that the energy-conditioned noise schedule also affects contrast and chrominance.

\subsection{Discussion}

The results distinguish the roles of the energy-conditioned components. Global spectral whitening provides dataset-level normalization while preserving image-dependent energy variations. The effect of $\psi_k$ is primarily learned during training: using the ECSdiff-trained model with the EAFW sampling configuration confirms that it need not be reapplied during sampling. Setting $\psi_k=1$ improves FID at almost all evaluated checkpoints, reaching $100.45$ with $10000$ samples compared with $106.55$ when $\psi_k$ is retained. The variability observed with $1000$ samples (Figure~\ref{fig4}) is therefore largely attributed to the limited FID sample size. In contrast, the image-dependent $\beta_E^{(i)}(t)$ path remains relevant for energy-conditioned training and potentially provides a more important conditioning mechanism during sampling.

The gradual introduction of higher spectral weights $\omega$ over diffusion steps $t$ and training epochs $e$~\eqref{eq-white}, together with the epoch-dependent learning rate $\mathrm{lr}(e)$, improves optimization and accelerates convergence during the early training epochs.

Future experiments will extend ECSdiff to transformer-based architectures and larger training datasets with higher-resolution images, such as ImageNet at $256\times256$ resolution, to examine whether the improvements demonstrated here with the compact DCTdiff architecture are preserved across different spectral diffusion models.

\section*{Acknowledgments}
The authors would like to thank OpenAI’s ChatGPT \cite{Openai2025chatgpt} for assistance with language editing, translation, and improving text clarity during manuscript preparation.

\balance
\bibliographystyle{model1-num-names}
\bibliography{cas-refs.bib}


\bio{}
\textbf{Bata Vasic} received the B.Sc. and M.Sc. degrees in electrical engineering and telecommunications and the Ph.D. degree in electrical engineering and computer science from the University of Niš, Serbia. He is a Research Professor at the University of Niš and an Associate Professor at MB University, Belgrade, where he teaches courses in computer graphics, artificial intelligence, cryptography, and watermarking. He has served as a Visiting Professor at ENSEA, University of Cergy-Pontoise, France, and leads EU-funded research projects on AI development and applications. His research interests include neural network architectures, generative AI security, authorship protection and provenance in diffusion models, digital watermarking, and 3D mesh geometry.
\endbio

\bio{}
\textbf{Bane Vasic} received the B.Sc., M.Sc., and Ph.D. degrees from the University of Niš, Serbia. He is a Professor of Electrical and Computer Engineering and Mathematics at the University of Arizona and Director of the Error Correction Laboratory. His research has made significant contributions to error correction, including structured low-density parity-check codes and decoding algorithms for data storage systems. He leads research on quantum error correction within major U.S. Department of Energy, NASA, and National Science Foundation programs. He is an IEEE Fellow, Fulbright Scholar, and da Vinci Fellow, and a co-founder of Codelucida and QEC Labs. His research interests include coding theory, classical and quantum error correction, iterative decoding, and reliable data storage and quantum computing systems.
\endbio

\end{document}